\documentclass[journal, table, xcdraw, twoside]{IEEEtran}
\IEEEoverridecommandlockouts
\usepackage{adjustbox}
\usepackage{amsmath,amssymb,amsfonts}
\usepackage{algorithmic}
\usepackage{color}
\usepackage{environ}
\usepackage{float}
\usepackage{graphicx}
\usepackage[utf8]{inputenc}
\usepackage{lipsum}  
\usepackage{listings}
\usepackage{multirow}
\usepackage[numbers, square, comma,compress]{natbib}
\usepackage{pgfplots}
\usepackage{relsize}
\usepackage{siunitx}
\usepackage{subcaption}
\usepackage{textcomp}
\usepackage{tikz}
\usepackage{tikz-3dplot}
\usepackage{balance}
\usepackage{url}
\usepackage{fancyhdr}

\fancypagestyle{firststyle}
{
   \fancyhf{}
   \fancyhead[R]{\footnotesize \thepage}
   \fancyhead[L]{\footnotesize IEEE ROBOTICS AND AUTOMATION LETTERS. PREPRINT VERSION. ACCEPTED FEBRUARY, 2021}
}

\DeclareUnicodeCharacter{2212}{−}
\usepgfplotslibrary{groupplots,dateplot}
\usetikzlibrary{patterns,shapes.arrows, matrix}
\pgfplotsset{compat=newest}

\NewDocumentCommand{\evalat}{sO{\big}mm}{%
  \IfBooleanTF{#1}
   {\mleft. #3 \mright|_{#4}}
   {#3#2|_{#4}}%
}

\newcommand{\gpmpc}{GP-MPC}

\newcommand{\vect}[3]{{_{\mathsmaller{\mathrm{#2}}}\mathbf{#1}_{\mathsmaller{\mathrm{#3}}}}} %
\newcommand{\wfr}[0]{\ensuremath{W}} %
\newcommand{\bfr}[0]{\ensuremath{B}} %

\newcommand{\bm}[1]{\boldsymbol{#1}}
\newcommand{\mat}[1]{\begin{bmatrix}#1\end{bmatrix}}
\newcommand{\rom}[1]{(\expandafter{\romannumeral #1\relax})}

\newcommand{\comment}[1]{}
\newcommand{\TODO}[1]{{\color{blue}{TODO: #1}}}
\newcommand{\rebuttal}[1]{{\color{black}{#1}}}

\begin{document}

\newsavebox\mybox
\newenvironment{resizedtikzpicture}[1]{%
  \def\mywidth{#1}%
  \begin{lrbox}{\mybox}%
  \begin{tikzpicture}
}{%
  \end{tikzpicture}%
  \end{lrbox}%
  \resizebox{\mywidth}{!}{\usebox\mybox}%
}

\title{
Data-Driven MPC for Quadrotors
}

\author{Guillem Torrente$^{\ast}$, %
        Elia Kaufmann$^{\ast}$,        %
        Philipp F\"{o}hn,
        Davide Scaramuzza
        \thanks{
        Manuscript received: October 15th, 2020; Revised: December 26th, 2020; Accepted: February 2nd, 2021}
        \thanks{
        This paper was recommended for publication by Editor Clement Gosselin upon evaluation of the Associate Editor and Reviewers’ comments.}
        \thanks{    
        $^{\ast}$These two authors contributed equally.}
        \thanks{The authors are with the Robotics and Perception Group, Dep. of Informatics, University of Zurich, and Dep. of Neuroinformatics, University of Zurich and ETH Zurich, Switzerland (\protect\url{http://rpg.ifi.uzh.ch}).
        This work was supported by the National Centre of Competence in Research (NCCR) Robotics through the Swiss National Science Foundation (SNSF) and the European Union’s Horizon 2020 Research and Innovation Programme under grant agreement No. 871479 (AERIAL-CORE) and the European Research Council (ERC) under grant agreement No. 864042 (AGILEFLIGHT).
        }
        \thanks{Digital Object Identifier (DOI): see top of this page.}
        }

\maketitle
\thispagestyle{firststyle}

\begin{abstract}

Aerodynamic forces render accurate high-speed trajectory tracking with quadrotors extremely challenging.
These complex aerodynamic effects become a significant disturbance at high speeds, introducing large positional tracking errors, and are extremely difficult to model.
To fly at high speeds, feedback control must be able to account for these aerodynamic effects in real-time.
This necessitates a modeling procedure that is both accurate and efficient to evaluate.
Therefore, we present an approach to model aerodynamic effects using Gaussian Processes, which we incorporate into a Model Predictive Controller to achieve efficient and precise real-time feedback control, leading to up to 70\% reduction in trajectory tracking error at high speeds.
We verify our method by extensive comparison to a state-of-the-art linear drag model in synthetic and real-world experiments at speeds of up to 14m/s and accelerations beyond 4g.

\end{abstract}

\section*{Supplementary Material}
\rebuttal{
\noindent Video: \url{https://youtu.be/FHvDghUUQtc}\\
\noindent Code: \url{https://github.com/uzh-rpg/data\_driven\_mpc}
}

\section{Introduction}\label{sec:introduction}

Accurate trajectory tracking with quadrotors in high-speed and high-acceleration regimes is still a challenging research problem. 
While autonomous quadrotors have seen a significant gain in popularity and have been applied in a variety of industries ranging from agriculture to transport, security, infrastructure, entertainment, and search and rescue, they still do not exploit their full maneuverability. 
The ability to precisely control drones during fast and highly agile maneuvers
would allow to not only fly fast in known-free environments, but also close to obstacles, humans, or through openings, where already small deviations from the reference have catastrophic consequences. 

Operating a quadrotor at high speeds and controlling it through agile, high-acceleration maneuvers requires to account for complex aerodynamic effects acting on the platform. 
These effects are difficult to model, since they consist of a combination of propeller lift and drag dependent on the induced airstream velocity, fuselage drag, and complex or even turbulent effects due to the interaction between the propellers, the downwash of other propellers, and the fuselage.
Furthermore, in the context of model-based feedback control, the model complexity is constrained by the feedback time-scale and computational capabilities of the executing platform.
Therefore, it is not sufficient to find the most accurate model, but required to find an applicable trade-off between model accuracy and complexity.

Very little work exists on agile control of quadrotors at speeds beyond $\SI{5}{\meter\per\second}$ and accelerations above 2g,~\cite{falanga2018pampc, faessler2015automatic, kamelmpc2016, Neunert16icra, mellinger2012trajectory, kaufmann2020RSS, foehn2020alphapilot, yunlong2020learning}.
Even though these works show agile control at various levels, none of them accounts for aerodynamic effects. 
This is not a limiting assumption when the quadrotor is controlled close to hover conditions, but introduces significant errors when tracking fast and agile trajectories.
Other approaches use iterative learning control to perform highly aggressive trajectories~\cite{lupashin2010simple}, but they are constrained to a single maneuver and do not generalize.

\begin{figure}[t]
    \centering
    \includegraphics[width=1.0\linewidth]{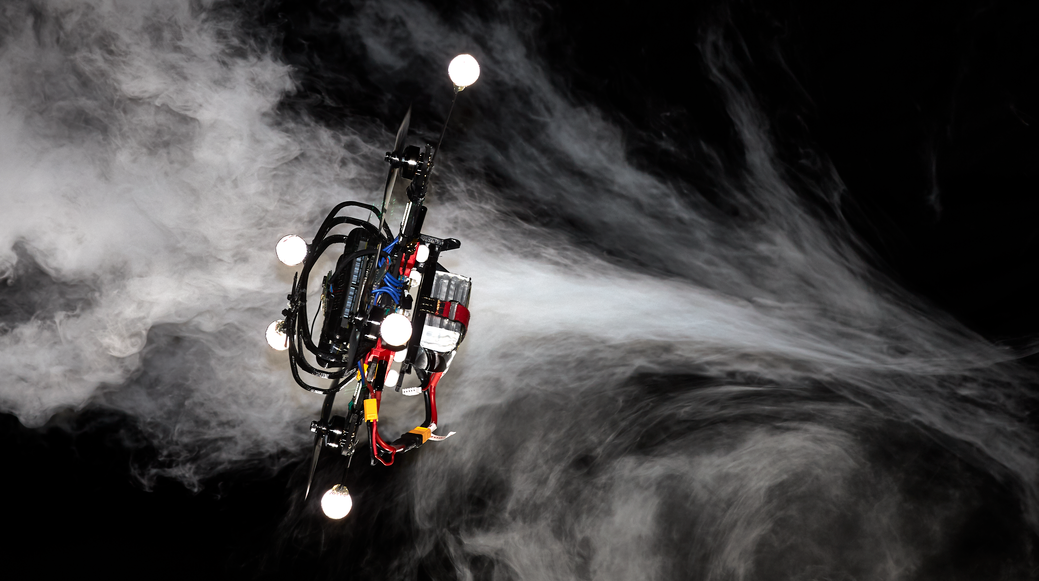}
    \caption{Our quadrotor platform reaches its physical limits at a pitch angle of $80$ degrees while performing a lemniscate trajectory in our experiments.  Throughout the trajectory, the platform reaches speeds of up to $14\si{\meter\per\second}$ and accelerations beyond 4g.}
    \label{fig:catch-eye}
\end{figure}

The main challenge when performing aggressive flight is to identify a dynamics model of the platform that is capable of describing the aerodynamic effects while still being lightweight enough to guarantee real time performance.
While there exist sophisticated computational fluid dynamics simulations that are able to model turbulent aerodynamic effects~\cite{Diaz2018aiaa}, they require hours of processing on a compute cluster, and still need to be abstracted in simplified models to be tractable in a control loop running at high frequency. 

In this work, we propose to learn the aerodynamic effects acting on the platform from data. 
Inspired by~\cite{hewing2019cautious, kabzan2019learning}, we use Gaussian Processes to learn the residual dynamics with respect to a simplified quadrotor model that does not account for aerodynamic effects. 
Learning the residual dynamics simplifies the learning problem and allows describing the model augmentation using only a small number of inducing points for Gaussian Processes.
Using such a small model allows leveraging the combined dynamics formulation in a Model Predictive Control (MPC) pipeline.

Our experiments, performed in simulation and in the real world, show that the proposed approach can significantly improve control performance for agile trajectories with speeds up to $\SI{14}{\meter\per\second}$ and accelerations exceeding 4g. 
We show that the method generalizes between different trajectories and outperforms methods relying on simplified correction terms.

\subsection*{Contributions}
In this paper, we present a Model Predictive Control pipeline that is augmented with learned residual dynamics using Gaussian Processes. 
We extend the approach presented in~\cite{hewing2019cautious, kabzan2019learning} to three-dimensional GP predictions for the quadrotor platform.
By combining the learned GP corrections with the nominal quadrotor dynamics, we can learn an accurate dynamics model from a small number of inducing data points. 
\rebuttal{Such a small model can be efficiently optimized within an MPC pipeline and allows for control frequencies greater than \SI{100}{\hertz}.
}
We show that the augmented MPC improves trajectory tracking by up to 70\% with respect to its nominal counterpart. 
We verify our method by extensive comparison to a state-of-the-art linear drag model in synthetic and real-world experiments at speeds of up to $14\si{\meter\per\second}$ and accelerations beyond 4g.

\section {Related Work}

Performing fast and agile maneuvers with an autonomous robot requires knowledge of an accurate dynamics model of the platform. 
However, especially at high speeds and accelerations, such a description of the system is difficult to obtain due to hard-to-model effects caused by friction, aerodynamics or varying battery voltage. 
As modeling these effects significantly increases the problem complexity, controlling a robot under such conditions requires to find a trade-off between model expressivity and computational tractability. 
Most prior work on control of autonomous robots does not account for higher-order effects at all and treats them as external disturbances~\cite{falanga2018pampc, bicego2020nonlinear, liu2015explicit, ru2017nonlinear, lunni2017nonlinear}.
While this allows for very efficient and lightweight controller implementations, tracking performance progressively decreases for higher speeds. 
\rebuttal{In~\cite{tal2020accurate}, Nonlinear Incremental Dynamics Inversion is used to achieve robust tracking of fast trajectories. 
However, the reactive nature of this approach does not allow to account for future disturbances as the controller does not optimize over a horizon of actions. 
}

A recent line of work~\cite{williams2017information, fan2020deep, bieker2020deep, lenz2015deepmpc} investigates the application of dynamics learned entirely from data for a variety of applications such as robot arms, cars or fluids.
These learned dynamics representations take the form of deep neural networks and substitute the nominal dynamics in the MPC. 
While the resulting dynamics models are very expressive, their optimization is often intractable due to local minima. 
A common way to overcome this challenge is to use sampling-based optimizers, which in turn scale poorly to high-dimensional input spaces. 

Instead of learning the full dynamics from data, \cite{hewing2019cautious, kabzan2019learning, saveriano2017data, carron2019data} combine a nominal model with a learned correction term. This allows to limit the learned dynamics to have different dimensionality than the nominal system and provides the possibility to learn only specific effects that are difficult to capture with the nominal model. 

For the particular case of quadrotor flight, the most prominent source of disturbances are aerodynamic effects originating from drag caused by the rotors and the fuselage, as well as lift effects that act on the platform at high speeds. 
By conducting controlled experiments in both wind tunnels as well as instrumented tracking volumes, previous work has shown a significant effect of aerodynamic forces already at linear speeds of $\SI{5}{\meter\per\second}$~\cite{faessler2017differential, sun2019quadrotor}.

Neglecting other aerodynamic effects, previous work mainly studies the effect of rotor drag~\cite{bristeau2009role, faessler2017differential, martin2010true}. 
Rotor drag effects originate from blade flapping and induced drag of the rotors. These effects are typically combined as linear effects in a lumped parameter dynamical model~\cite{mahony2012multirotor}.
In~\cite{faessler2017differential}, the authors identify a linear model for the rotor drag and use it to compute feedforward terms of a PID controller. 
Even though they show improved trajectory tracking performance, they evaluate their linear model only up to $\SI{5}{\meter\per\second}$. 
At these speeds, the linear effect of rotor drag dominates the fuselage drag. 
We integrate the model of~\cite{faessler2017differential} in an MPC pipeline to act as baseline of our approach.

Similar to our work, in~\cite{hewing2019cautious, kabzan2019learning, mehndiratta2020gaussian, cao2017gaussian, desaraju2018leveraging}, the authors use Gaussian Processes to improve the control performance of a robotic platform.
In~\cite{mehndiratta2020gaussian}, Gaussian Processes are used on a quadrotor to correct for wind disturbances. 
Since instead of platform states only observed disturbances are fed to the GPs, this approach does not learn a dynamics model and can only react to disturbances once they have been observed.
In~\cite{cao2017gaussian}, the authors separately learn the translational and rotational dynamics of a quadrotor platform. 
\rebuttal{As this approach learns the full model from data, it requires a large number of training points and is computationally very expensive. 
As a result, the prediction horizon of the MPC needs to be reduced to a single point to achieve near real-time performance.
}
\rebuttal{
In~\cite{desaraju2018leveraging}, a robust experience-driven predictive controller~(EPC) is proposed that uses Gaussian belief propagation to account for uncertainties in the state estimate. 
The controller demonstrates robust constraint satisfaction on a quadrotor platform, where it is integrated into an MPC that controls the translational dynamics of the vehicle.
}
In~\cite{hewing2019cautious, kabzan2019learning}, the authors use the predictions of the Gaussian Processes to improve the tracking performance of an autonomous race car by learning the \textit{residual} dynamics of a nominal model.
Learning such residual dynamics instead of the full model allows them to simplify the learning problem and as a result reduce the number of training points in the GP.

Our work is inspired by these approaches, but extends~\cite{hewing2019cautious, kabzan2019learning} to three-dimensional GP predictions for the quadrotor platform.
Instead of learning a mapping from observed disturbances to future disturbance 
as in~\cite{mehndiratta2020gaussian}, we focus on fast flight and correct for aerodynamic effects that arise due to the fast ego-motion of the platform.
We tightly integrate the predictions of the Gaussian Processes into the MPC formulation. Instead of using virtual inputs such as bodyrates and collective thrust~\cite{falanga2018pampc}, our MPC models the dynamics down to the motor inputs and can therefore account for the true actuation limits of the platform. 
\rebuttal{Our work is the first to combine Gaussian Processes with a full-state quadrotor MPC formulation to model aerodynamic drag effects while still being able to account for the true actuation limits of the platform.}

\section{Methodology}\label{sec:method}

\subsection{Notation}
\begin{figure}[t]
\centerline{
\resizebox{7.8cm}{!}{
    \tdplotsetmaincoords{65}{10}
\begin{tikzpicture}[tdplot_main_coords, scale=2]

\draw[very thick] (-1.4,0,1) -- (1.4,0,1);
\draw[very thick] (0,-1.4,1) -- (0,1.4,1);

\draw[draw=none, fill=gray!40, opacity=0.8] (0,1.4,1) circle (0.8) node[right] {3};
\draw[draw=none, fill=gray!40, opacity=0.8] (1.4,0,1) circle (0.8) node[right] {0};
\draw[draw=none, fill=gray!40, opacity=0.8] (0,-1.4,1) circle (0.8) node[right] {1};
\draw[draw=none, fill=gray!40, opacity=0.8] (-1.4,0,1) circle (0.8) node[left] {2};

\draw[very thick,->] (0,1.4,1) -- (0,1.4,1.5);
\draw[very thick,->] (1.4,0,1) -- (1.4,0,1.5);
\draw[very thick,->] (0,-1.4,1) -- (0,-1.4,1.5);
\draw[very thick,->] (-1.4,0,1) -- (-1.4,0,1.5);	

\draw[thick,->] (0,2.0,1) arc (90:0:0.6);
\draw[thick,<-] (2.0,0,1) arc (0:-90:0.6);
\draw[thick,->] (0.6,-1.4,1) arc (0:-90:0.6);
\draw[thick,->] (-2.0,0,1) arc (180:270:0.6);

\draw[thick,->,color=red,text=black] (0,0,1) -- (0.7071,0.7071,1) node[right] {$\vect{x}{}{\bfr}$};
\draw[thick,->,color=green,text=black] (0,0,1) -- (-0.7071,0.7071,1) node[above] {$\vect{y}{}{\bfr}$};	
\draw[thick,->,color=blue,text=black] (0,0,1) -- (0,0,2) node[left] {$\vect{z}{}{\bfr}$};		
\node[draw=none] at (0.2,0,0.85) {Body};	

\draw[thick,->,color=red,text=black] (-1.5,-1,-0.5) -- (-0.7,-1,-0.5) node[right] {$\vect{x}{}{\wfr}$};		
\draw[thick,->,color=green,text=black] (-1.5,-1,-0.5) -- (-1.5,-0.2,-0.5) node[right] {$\vect{y}{}{\wfr}$};		
\draw[thick,->,color=blue,text=black] (-1.5,-1,-0.5) -- (-1.5,-1,0.3) node[above] {$\vect{z}{}{\wfr}$}; %
\node[draw=none] at (-1.6,0,-1.1) {World};

\draw[very thick,->,color=black,text=black] (2,-1.0,0.8) -- node[right] {$\bm{g}\vect{}{}{\wfr}$} (2,-1.0,0);

\end{tikzpicture}
}}
\caption{Diagram of the quadrotor model with the world and body frames and propeller numbering convention.}
\label{fig:quad_top_down_diagram}
\end{figure}
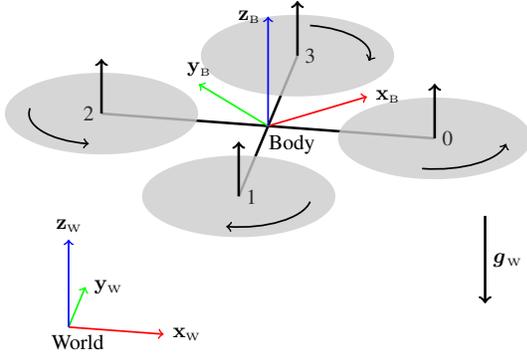

We denote scalars in lowercase $s$, vectors in lowercase bold $\bm{v}$, and matrices in uppercase bold $\bm{M}$.
We define the World $W$ and Body $B$ frames with orthonormal basis i.e. $\{\bm{x}_W, \bm{y}_W, \bm{z}_W\}$.
The frame $B$ is located at the center of mass of the quadrotor.
Note that we assume all four rotors are situated in the $xy$-plane of frame $B$, as depicted in Fig.~\ref{fig:quad_top_down_diagram}.
A vector from coordinate $\bm{p}_1$ to $\bm{p}_2$ expressed in the $W$ frame is written as: $_W\bm{v}_{12}$.
If the vector's origin coincide with the frame it is described in, we drop the frame index, e.g. the quadrotor position is denoted as $\bm{p}_{WB}$.
Furthermore, we use unit quaternions $\bm{q} = (q_w, q_x, q_y, q_z)$ with $\|\bm{q}\| = 1$ to represent orientations, such as the attitude state of the quadrotor body $\bm{q}_{WB}$.
Finally, full SE3 transformations, such as changing the frame of reference from body to world for a point $\bm{p}_{B1}$, can be described by $_W\bm{p}_{B1} = _W\bm{t}_{WB} + \bm{q}_{WB} \odot \bm{p}_{B1}$.
Note the quaternion-vector product denoted by $\odot$ representing a rotation of the vector by the quaternion as in $\bm{q} \odot \bm{v} = \bm{q} \bm{v} \bar{\bm{q}}$, where $\bar{\bm{q}}$ is the quaternion's conjugate.

\subsection{Nominal Quadrotor Dynamics Model}
We assume the quadrotor is a 6 degree-of-freedom rigid body of mass $m$ and diagonal moment of inertia matrix $\bm{J}=\mathrm{diag}(J_x, J_y, J_z)$.
Our model is similar to~\cite{falanga2018pampc, kamelmpc2016} but we write the nominal dynamics $\dot{\bm{x}}$ up to second order derivatives, leaving the quadrotors individual rotor thrusts $T_i \forall i \in (0, 3)$ as control inputs $\bm{u}$. 
The state space is thus 13-dimensional and its dynamics can be written as:
\begin{align}
\small
\label{eq:3d_quad_dynamics}
\dot{\bm{x}} =
\begin{bmatrix}
\dot{\bm{p}}_{WB} \\  
\dot{\bm{q}}_{WB} \\
\dot{\bm{v}}_{WB} \\
\dot{\boldsymbol\omega}_B
\end{bmatrix} = 
\bm{f}_{dyn}(\bm{x}, \bm{u}) =
\begin{bmatrix}
\bm{v}_W \\  
\bm{q}_{WB} \cdot \mat{0 \\ \bm{\omega_B}/2} \\
\frac{1}{m}\;\bm{q}_{WB} \odot \bm{T}_B + \bm{g}_W \\
\bm{J}^{-1}\left(\boldsymbol{\tau}_B - \boldsymbol\omega_B \times \bm{J}\boldsymbol\omega_B\right)
\end{bmatrix} \; ,
\end{align}
where $\bm{g}_W= [0, 0, \SI{-9.81}{\meter\per\second^2}]^\intercal$ denotes Earth's gravity, $\bm{T}_B$ is the collective thrust and $\bm{\tau}_B$ is the body torque as in:
\begin{align}
\small
\bm{T}_B &= \! \mat{0 \\ 0 \\ \sum T_i} &
\!\!\text{and}\!\! \quad 
\bm{\tau}_B &= \!
\mat{d_y (-T_0 - T_1 + T_2 + T_3) \\
d_x (-T_0 + T_1 + T_2 - T_3) \\
c_\tau (-T_0 + T_1 - T_2 + T_3)}
\end{align}
where $d_x$, $d_y$ are the rotor displacements and $c_\tau$ is the rotor drag torque constant.
To incorporate these dynamics in discrete time algorithms, we use an explicit Runge-Kutta method of 4th order $\bm{f}_{RK4}(\bm{x}, \bm{u})$ to integrate $\dot{\bm{x}}$ given an initial state $\bm{x}_k$, input $\bm{u}_k$ and integration step \rebuttal{$\delta t$} by:
\begin{equation}
\label{eq:discretized_nominal_dynamics}
\rebuttal{\bm{x}_{k+1} = \bm{f}_{RK4}(\bm{x}_k, \bm{u}_k, \delta t).}
\end{equation}

\subsection{Gaussian Process-Augmented Dynamics}
Inspired by~\cite{hewing2019cautious, kabzan2019learning}, we use Gaussian Processes to complement the nominal dynamics of the quadrotor in an MPC pipeline.
In this setting, the GPs predict the error of the dynamics and correct them at every time instance $t_k$. 
Similar to most GP-based learning problems, we assume the existence of the inaccessible true dynamics \rebuttal{$\bm{f}_{true}$} of the quadrotor, which we measure as $\tilde{\bm{y}}_{k+1}$ through a noisy process at discrete time instances $t_k$:
\begin{align}
    \tilde{\bm{y}}_{k+1} &= \bm{f}_{true}(\bm{x}_k, \bm{u}_k) + \bm{w}_k
\end{align}
We further assume that $\bm{w}_k\sim\mathcal{N}(\bm{0}, \bm{\Sigma})$ is Gaussian noise, where $\bm{\Sigma}$ is the time-invariant and diagonal covariance matrix.
This means we can effectively treat each dimension of $\bm{\bm{y}}_k$ independently through a separate 1-dimensional output GP. 
We use the Radial Basis Function (RBF) kernel defined by:
\begin{align}\label{eq:rbf}
\kappa(\bm{z}_i, \bm{z}_j) = \sigma_f^2 \exp\left(-\frac{1}{2}
  (\bm{z}_i - \bm{z}_j)^\intercal \bm{L}^{-2}
  (\bm{z}_i - \bm{z}_j)\right) + \sigma_n^2 %
\end{align}
where $\bm{L}$ is the diagonal length scale matrix and $\sigma_f$, $\sigma_n$ represent the data and prior noise variance, respectively, and $\bm{z}_i$, $\bm{z}_j$ represent data features.

We redefine the system dynamics as a (corrected\rebuttal{, $\bm{f}_{corr}$}) combination of the dynamics \eqref{eq:3d_quad_dynamics} plus the mean posterior of a GP, $\bm{\mu}$.
The GP only corrects a subset of the state, determined in the selection matrix $\bm{B}_d$, using the feature vector $\bm{z}_k$, determined by selection matrix $\bm{B}_z$:
\begin{align}
\rebuttal{\bm{f}_{cor}(\bm{x}_k, \bm{u}_k)} &= \bm{f}_{dyn}(\bm{x}_k, \bm{u}_k) + \bm{B}_d \bm{\mu}(\bm{z}_k) \\
\bm{z}_k &= \bm{B}_z \mat{ \bm{x}_k^\intercal & \bm{u}_k^\intercal}^\intercal.
\end{align}

Given the concatenated training feature samples $\bm{Z}$ and the query feature samples $\bm{Z}_k$, the mean and covariance of the GP prediction can be recovered as follows:
\begin{align}
\bm{\mu}(\bm{Z}_k) &= \bm{K}_k^\intercal \bm{K}^{-1} \bm{Z}&
\bm{\Sigma}_{\mu k} &= \bm{K}_{kk} - \bm{K}_k^\intercal \bm{K}^{-1} \bm{K}_k \nonumber \\
\text{with } \quad
\bm{K} &= \kappa(\bm{Z}, \bm{Z}) + \sigma_n^2 \bm{I} && \\
\bm{K}_k &= \kappa(\bm{Z}, \bm{Z}_k) &
\bm{K}_{kk} &= \kappa(\bm{Z}_k, \bm{Z}_k). \nonumber
\end{align} 
where $\bm{K}_{ij}$, the entry of $\bm{K}$ with index $i, j$, is $\bm{K}_{ij} = \kappa(\bm{z}_i, \bm{z}_j)$.

Given the mean and covariance of the GP, not only is it possible to learn the corrected dynamics, but in addition we can also propagate the corrected model forward in time to use it in an MPC.
To propagate the state we simply substitute the nominal dynamics $\bm{f}_{dyn}$ with the corrected model $\bm{f}_{cor}$ in the Runge-Kutta integration.
\rebuttal{For the propagation of the covariance, we refer to the formulation in~\cite{hewing2019cautious, kabzan2019learning}.}

\subsection{MPC Formulation}

In its most general form, MPC stabilizes a system subject to its dynamics $\dot{\bm{x}} = \bm{f}(\bm{x}, \bm{u})$ along a reference ${\bm{x}^*(t), \bm{u}^*(t)}$, by minimizing a cost $\mathcal{L}(\bm{x}, \bm{u})$ as in:

\begin{align}
\min_{\bm{u}} &\int \mathcal{L}(\bm{x}, \bm{u}) \\
\text{subject to } \quad 
\quad \dot{\bm{x}} &= \bm{f}_{dyn}(\bm{x}, \bm{u}) &
\bm{x}(t_0) &= \bm{x}_{init} \nonumber \\
\bm{r}(\bm{x}, \bm{u}) &= 0 & 
\bm{h}(\bm{x}, \bm{u}) &\leq 0 \nonumber
\end{align}
where $\bm{x}_0$ denotes the initial condition and $\bm{h}$, $\bm{r}$ can incorporate (in-)equality constraints, such as input limitations.

For our application, and as most commonly done, we specify the cost to be of quadratic form $\mathcal{L}(\bm{x}, \bm{u}) = \| \bm{x} - \bm{x}^*  \|^2_Q + \| \bm{u} - \bm{u}^* \|^2_R$ and discretize the system into $N$ steps over time horizon $T$ of size $dt=T/N$.
We account for input limitations by constraining $0 \leq \bm{u} \leq u_{max}$, and optionally include the GP predictions within the system dynamics.

\begin{align}
\min_{u} \bm{x}_N^\intercal Q \bm{x}_N + &\sum_{k=0}^N \bm{x}_k^\intercal Q \bm{x}_k + \bm{u}_k^\intercal R \bm{u}_k \\
\text{subject to} \quad
\bm{x}_{k+1} &= \rebuttal{\bm{f}_{RK4}(\bm{x}_k, \bm{u}_k, \delta t)} \nonumber \\
\bm{x}_0 &= \bm{x}_{init} \nonumber \\
u_{min} & \leq \bm{u}_k \leq u_{max} \nonumber 
\end{align}
\rebuttal{where $\bm{f}_{RK4}$ can be extended to the corrected dynamics $\bm{f}_{cor}$}.

To solve this quadratic optimization problem we construct it using a multiple shooting scheme \cite{Diehl2006springer} and solve it through a sequential quadratic program (SQP) executed in a real-time iteration scheme (RTI) \cite{Diehl2006springer}.
All implementations are done using ACADOS~\cite{acados} and CasADi~\cite{Andersson2019}.

\subsection{Practical Implementation}

The implementation of the learned dynamics of our GP-MPC must be designed to maximize the performance while minimizing the computational cost added to the optimization.
Note that aerodynamic effects operate on the body reference frame.
Likewise, the training dataset is adjusted such that the learning problem setup is to identify such a mapping from body frame velocities $_B\bm{v}$ to body frame acceleration disturbances $_B\bm{a}_e$, so that $_B\bm{a}_e = \bm{\mu}(_B\bm{v})$.
Furthermore, to reduce the need for additional training samples, we reduce the dimensionality of our input space such that the mappings are learned axis wise.  
We can thus write the GP prediction as:
\begin{align}
\label{eq:gp_inference}
_B\bm{a}_{ek} = \bm{\mu}(_B\bm{v}_k) &=
\mat{
\mu_{vx}(_Bv_{xk}) \\
\mu_{vy}(_Bv_{yk}) \\
\mu_{vz}(_Bv_{zk})
} \\
\bm{\Sigma}_\mu(_B\bm{v}_k) &= \mathrm{diag}\left(
\mat{
\sigma_{vx}^2(_Bv_{xk}) \\
\sigma_{vy}^2(_Bv_{yk}) \\
\sigma_{vz}^2(_Bv_{zk})}\right)
\end{align}

\subsection{Data Collection and Model Learning}

To fit the GPs, real-world flight data is collected (as detailed in Sec. \ref{sec:experiments}) using the nominal dynamics model.
For each sample at time $t_k$, the velocity at the next sample point $_B\bm{v}_{k+1}$ and the predicted velocity at the next sample point $_B\hat{\bm{v}}_{k+1}$ \rebuttal{are} recorded, together with the timestep $\delta t_k$.
We can then compute the time-normalized velocity error, corresponding to the acceleration error:
\rebuttal{
\begin{equation}
_B\bm{a}_{ek} = \frac{_B\bm{v}_{k+1} - _B\hat{\bm{v}}_{k+1}}{\delta t_k}
\end{equation}
}

To select the hyperparameters of the kernel function $(l, \sigma_n, \sigma_f)$, we perform maximum likelihood optimization on the collected dataset. 
Being a non-parametric method, the complexity of GP regression depends on the number of training points. 
As using the full dataset would make MPC optimization intractable in real time, we subsample the dataset and only use a small number of inducing points. 
To this end, we leverage the smooth nature of the aerodynamic effects by sampling these points at regular intervals in the ranges of the training set.

\section{Experiments and Results}\label{sec:experiments}

We design our evaluation procedure to address the following questions:
\rom{1} What is the contribution of the learned dynamics of our GP-MPC in a closed-loop tracking task?
\rom{2} How does our GP-MPC compare to an MPC with linear aerodynamic effect compensation, as proposed in \cite{faessler2017differential}?
\rom{3} How does the learned model generalize to unseen trajectories?
Finally, we validate our design choices with ablation studies.
We refer the reader to the attached video to understand the dynamic nature of our experiments. 

\begin{figure*}[h]
\begin{adjustbox}{width=\textwidth}
\centerline{\includegraphics[width=\textwidth]{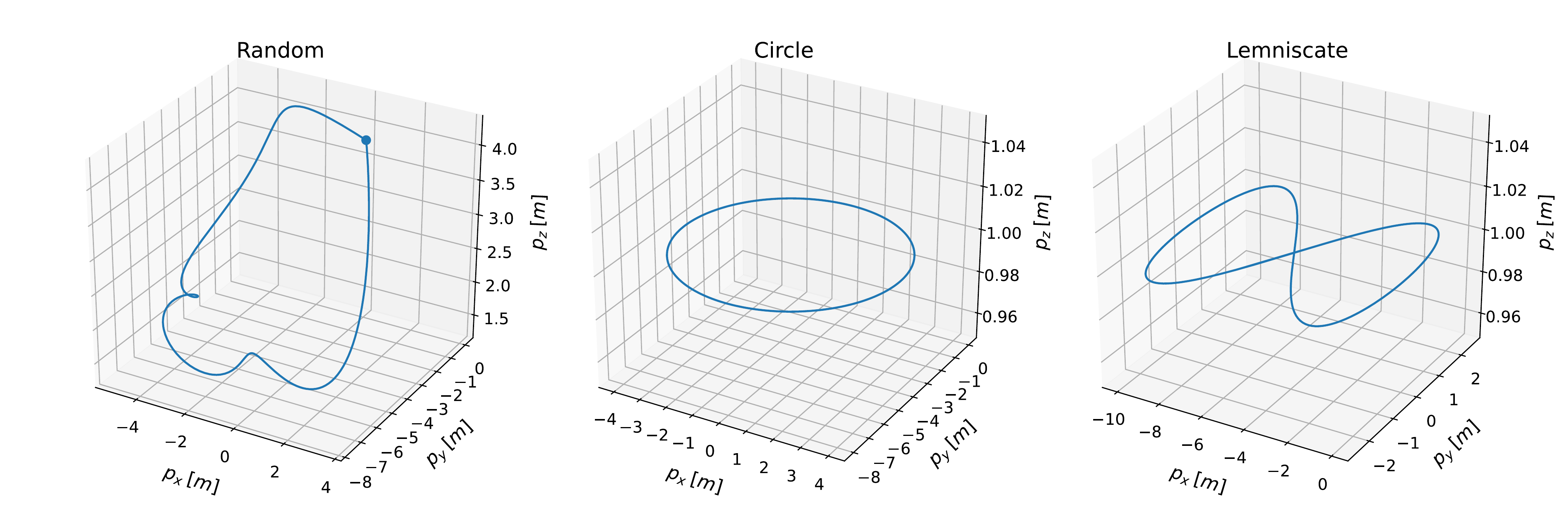}}
\end{adjustbox}
\caption{Trajectories considered in this work. Left: single-loop randomly-generated polynomial trajectory with motion along all axes. The tracking starts and ends at the upper-right corner. Center and right: circular and lemniscate trajectories respectively. Both have zero translation along the z axis, tracking starts at 0 velocity, ramps up until reaching a peak, and ramps down back to hover. The position references remain as shown in the figures in all cases.}
\label{fig:trajectory_plots}
\end{figure*}

\begin{figure*}
\begin{adjustbox}{width=\textwidth}%
\begin{tikzpicture}

\definecolor{color0}{rgb}{0.12156862745098,0.466666666666667,0.705882352941177}
\definecolor{color1}{rgb}{1,0.498039215686275,0.0549019607843137}
\definecolor{color2}{rgb}{0.172549019607843,0.627450980392157,0.172549019607843}
\definecolor{color3}{rgb}{0.83921568627451,0.152941176470588,0.156862745098039}

\begin{groupplot}[group style={group size=3 by 1, horizontal sep=1.3cm}, width=9cm, height=6cm]
\nextgroupplot[
tick align=outside,
tick pos=left,
title={Random},
x grid style={white!69.0196078431373!black},
xlabel={$v_{max}\:[\si{\meter\per\second}]$},
xmajorgrids,
xmin=7.42529616559266, xmax=14.0026795503627,
xtick style={color=black},
y grid style={white!69.0196078431373!black},
ylabel={RMSE $[\si{\meter}]$},
ymajorgrids,
ymin=-0.00709345229633279, ymax=0.192663010425758,
ytick style={color=black},
ytick={0,0.05,0.1,0.15,0.2},
yticklabels={0,0.05,0.1,0.15,0.2}
]
\addplot [ultra thick, color0, dashed, mark=*, mark size=3, mark options={solid}]
table {%
7.72426813762767 0.00198638691830772
10.5462718053015 0.00698995254737409
11.663798922937 0.0120449379606475
12.4642619710369 0.0154456459195535
13.7037075783277 0.0209206569533825
};
\coordinate (top) at (rel axis cs:0,1);%

\addplot [semithick, color1, dashed, mark=*, mark size=3, mark options={solid}]
table {%
7.72426813762767 0.124916562935251
10.5462718053015 0.163008049701881
11.663798922937 0.166967074332722
12.4642619710369 0.165918245343755
13.7037075783277 0.183510281152514
};
\addplot [semithick, color2, dashed, mark=*, mark size=3, mark options={solid}]
table {%
7.72426813762767 0.0423088459782289
10.5462718053015 0.0433836375206533
11.663798922937 0.0455569641980606
12.4642619710369 0.0451122090548598
13.7037075783277 0.0479068103589532
};
\addplot [semithick, color3, dashed, mark=*, mark size=3, mark options={solid}]
table {%
7.72426813762767 0.0182036851265909
10.5462718053015 0.023341671591241
11.663798922937 0.0268587886830462
12.4642619710369 0.0285944627303829
13.7037075783277 0.0330863741175463
};

\nextgroupplot[
tick align=outside,
tick pos=left,
title={Circle},
x grid style={white!69.0196078431373!black},
xlabel={$v_{max}\:[\si{\meter\per\second}]$},
xmajorgrids,
xmin=1.60245501175827, xmax=13.3537917646522,
xtick style={color=black},
y grid style={white!69.0196078431373!black},
ymajorgrids,
ymin=-0.0165018250745117, ymax=0.349973755391383,
ytick style={color=black}
]
\addplot [ultra thick, color0, dashed, mark=*, mark size=3, mark options={solid}]
table {%
2.13660668234436 0.00160351873973954
4.80736503527482 0.000156155855756271
7.47812338820527 0.000487427215394814
10.1488817411357 0.00106745852115383
12.8196400940662 0.00191358039059544
};
\addplot [semithick, color1, dashed, mark=*, mark size=3, mark options={solid}]
table {%
2.13660668234436 0.0498697672654518
4.80736503527482 0.134237117415627
7.47812338820527 0.213113796239548
10.1488817411357 0.276807097445538
12.8196400940662 0.333884811248971
};
\addplot [semithick, color2, dashed, mark=*, mark size=3, mark options={solid}]
table {%
2.13660668234436 0.0477402481353983
4.80736503527482 0.0360572444592102
7.47812338820527 0.0417768169087441
10.1488817411357 0.0459292652606522
12.8196400940662 0.0447349867561495
};
\addplot [semithick, color3, dashed, mark=*, mark size=3, mark options={solid}]
table {%
2.13660668234436 0.0150178987866975
4.80736503527482 0.017019019076374
7.47812338820527 0.0226068998841799
10.1488817411357 0.0279061563739752
12.8196400940662 0.0292657277400758
};

\nextgroupplot[
tick align=outside,
tick pos=left,
title={Lemniscate},
x grid style={white!69.0196078431373!black},
xlabel={$v_{max}\:[\si{\meter\per\second}]$},
xmajorgrids,
xmin=2.13492967215171, xmax=18.8250637073041,
xtick style={color=black},
y grid style={white!69.0196078431373!black},
ymajorgrids,
ymin=-0.0147818218574284, ymax=0.311491562251468,
ytick style={color=black}
]
\addplot [ultra thick, color0, dashed, mark=*, mark size=3, mark options={solid}]
table {%
2.893572128295 4.87865111578198e-05
5.9172632332243 0.000707523804622657
10.500403772372 0.00316422611791936
13.9597424215952 0.0109270126114256
18.0664212511608 0.0279967580888971
};\label{plots:plot1}
\addplot [semithick, color1, dashed, mark=*, mark size=3, mark options={solid}]
table {%
2.893572128295 0.0499506172481581
5.9172632332243 0.124258998677527
10.500403772372 0.18701069835734
13.9597424215952 0.243843908121371
18.0664212511608 0.296678895687433
};\label{plots:plot2}
\addplot [semithick, color2, dashed, mark=*, mark size=3, mark options={solid}]
table {%
2.893572128295 0.0374569702195354
5.9172632332243 0.0353889821380341
10.500403772372 0.0394715512909917
13.9597424215952 0.0365757990459391
18.0664212511608 0.055621943492531
};\label{plots:plot3}
\addplot [semithick, color3, dashed, mark=*, mark size=3, mark options={solid}]
table {%
2.893572128295 0.0144727811750416
5.9172632332243 0.0169266699165708
10.500403772372 0.0226888202794869
13.9597424215952 0.026018751722704
18.0664212511608 0.0422267920451941
};\label{plots:plot4}
\coordinate (bot) at (rel axis cs:1,0);%
\end{groupplot}

\path (top|-current bounding box.north)--
      coordinate(legendpos)
      (bot|-current bounding box.north);
\matrix[
    matrix of nodes,
    anchor=south,
    draw,
    inner sep=0.2em,
    draw
  ]at([yshift=1ex]legendpos)
  {
    \ref{plots:plot1}& Ideal & [5pt]
    \ref{plots:plot2}& Nominal & [5pt]
    \ref{plots:plot3}& GP-MPC 15 & [5pt] 
    \ref{plots:plot4}& GP-MPC 100 \\};
    
\end{tikzpicture}
\end{adjustbox}
\caption{Closed-loop position tracking error as a function of maximum velocity achieved in our custom simulator. \emph{Ideal} denotes the nominal MPC performance in a disturbance-free scenario. \emph{Nominal} corresponds to the un-augmented MPC, and \textit{GP-MPC 15} and \textit{GP-MPC 100} are our GP-augmented controllers where the GP's have been trained with 15 and 100 training samples.}
\label{fig:experiment_1_custom_sim}
\end{figure*}
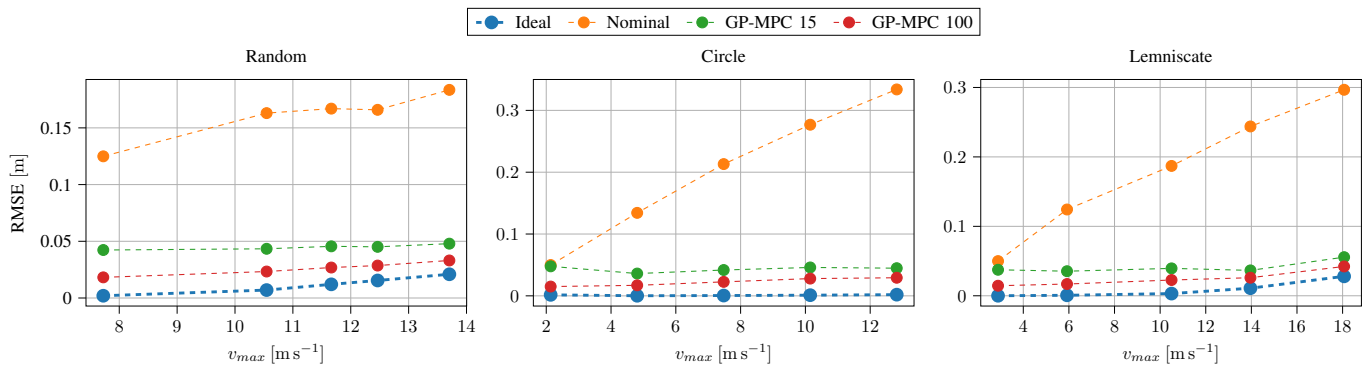

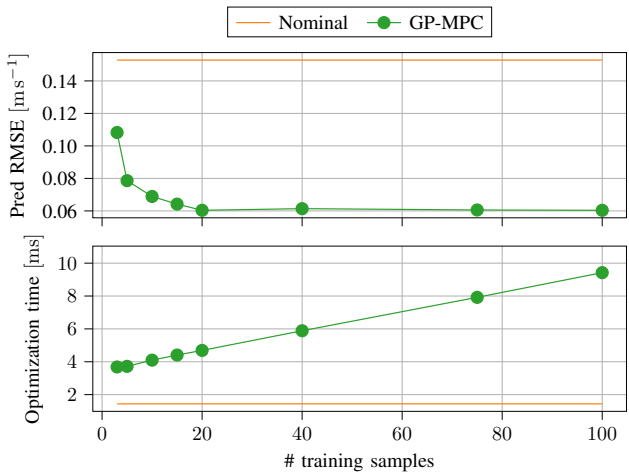
\begin{figure}[h] %
\begin{adjustbox}{width=0.95\columnwidth}%
\begin{tikzpicture}

\definecolor{color0}{rgb}{1,0.498039215686275,0.0549019607843137}
\definecolor{color1}{rgb}{0.172549019607843,0.627450980392157,0.172549019607843}

\begin{groupplot}[group style={group size=1 by 2, vertical sep=0.5cm},height=4.5cm,width=11cm]
\nextgroupplot[
tick align=outside,
tick pos=left,
x grid style={white!69.0196078431373!black},
xmajorgrids,
xmin=-1.85, xmax=104.85,
xtick style={color=black},
xticklabels={,,},
y grid style={white!69.0196078431373!black},
ylabel={Pred RMSE \([\si{\meter\per\second}]\)},
ymajorgrids,
ymin=0.0557551344209944, ymax=0.157436978728679,
ytick style={color=black},
ytick={0.06,0.08,0.10,0.12,0.14},
yticklabels={0.06,0.08,0.10,0.12,0.14}
]
\addplot [semithick, color0]
table {%
3 0.152815076714693
5 0.152815076714693
10 0.152815076714693
15 0.152815076714693
20 0.152815076714693
40 0.152815076714693
75 0.152815076714693
100 0.152815076714693
};
\coordinate (top) at (rel axis cs:0,1);%
\addplot [semithick, color1, mark=*, mark size=3, mark options={solid}]
table {%
3 0.10824897357623
5 0.0786098027252456
10 0.0688963705015471
15 0.0641781308024198
20 0.0603778055180326
40 0.0613776685258848
75 0.0605739911992048
100 0.06037703643498
};

\nextgroupplot[
tick align=outside,
tick pos=left,
x grid style={white!69.0196078431373!black},
xlabel={\# training samples},
xmajorgrids,
xmin=-1.85, xmax=104.85,
xtick style={color=black},
y grid style={white!69.0196078431373!black},
ylabel={Optimization time \([\si{\milli\second}]\)},
ymajorgrids,
ymin=0.982460228631913, ymax=11.0372916618834,
ytick style={color=black}
]
\addplot [semithick, color0]
table {%
3 1.43949802105243
5 1.43949802105243
10 1.43949802105243
15 1.43949802105243
20 1.43949802105243
40 1.43949802105243
75 1.43949802105243
100 1.43949802105243
};\label{tradeoff_plot:plot1}
\addplot [semithick, color1, mark=*, mark size=3, mark options={solid}]
table {%
3 3.68244965572966
5 3.72079613061157
10 4.09833754204899
15 4.40792742187042
20 4.6903389256175
40 5.88679358391109
75 7.91976331098889
100 9.42217028834359
};\label{tradeoff_plot:plot2}
\coordinate (bot) at (rel axis cs:1,0);
\end{groupplot}

\path (top|-current bounding box.north)--
      coordinate(legendpos)
      (bot|-current bounding box.north);
\matrix[
    matrix of nodes,
    anchor=south,
    draw,
    inner sep=0.2em,
    draw
  ]at([yshift=1ex]legendpos)
  {
    \ref{tradeoff_plot:plot1}& Nominal & [5pt]
    \ref{tradeoff_plot:plot2}& GP-MPC \\};
\end{tikzpicture}
\end{adjustbox}
\caption{Trade-off between number of training samples, GP performance (top) and optimization time (bottom). 
Models are trained and evaluated on data collected in our Simplified Simulation.}
\label{fig:gp_training_samples_tradeoff}
\end{figure}

\subsection{Experimental Setup}
We conduct experiments both in simulation as well as on a real quadrotor platform. 
To assess our proposed approach, the quadrotor executes three different trajectories (Random, Circle, Lemniscate), illustrated in Fig.~\ref{fig:trajectory_plots}. 
\rebuttal{The lemniscate trajectory lies in the horizontal plane and is defined by $\left[ x(t) = 2\cos\left( \sqrt{2}t\right); y(t) = 2 \sin\left( \sqrt{2}t\right)\cos\left(\sqrt{2}t\right)\right]$.}

We compare the tracking performance on these trajectories using our MPC with the \textit{Nominal} quadrotor model~\eqref{eq:3d_quad_dynamics}, and the improvement after adding different correction terms identified from data.
We study two possible augmentations: \textit{GP-MPC} (ours) and \textit{RDRv}. 
The RDRv approach was proposed in~\cite{faessler2017differential} as a feed-forward PID controller term, which identifies a set of linear drag coefficients along the body axes.
We instead incorporate this linear compensation into the nominal dynamics of our MPC pipeline.
Note that the three control approaches only differ in the dynamics model used by the MPC, i.e. they use the same control frequency and cost matrices.
\rebuttal{Furthermore, both the GP-MPC and the RDRv are always trained on the same dataset}.

For the simulation experiments, we perform a Nominal run on random polynomial trajectories of high aggressiveness to collect training samples for the GP and the coefficient identification for RDRv.
With both models fitted, we deploy all three controllers, Nominal, RDRv and GP-MPC, on the test trajectories without retraining.
For the real-world experiments, we first perform a run of the Nominal baseline on both circle and lemniscate trajectories, which is also used for GP training and RDRv coefficient identification.
In the subsequent rollouts, we test RDRv and GP-MPC on these two trajectories in different permutations.

\rebuttal{In our experimental setup, we only consider obstacle-free scenarios and aim to minimize the tracking error that arises mainly due to aerodynamic disturbance effects. %
For this reason, the predicted covariance of the GPs is not used in our experiments. 
}

\subsection{Experiments in Simulation}\label{sec:experiments_sim}
We first evaluate the performance for individual maneuvers in simulation. 
To isolate the effects of varying MPC computation times for different models, we divide the simulation experiments into two parts: \textit{simplified simulation} and \textit{Gazebo simulation}. 
This setup allows to compare the predictive performance of arbitrary sized models without the need to correct for varying computation times.

\textbf{Simplified Simulation} 
This simulation is constituted of a simple forward integration of the system dynamics~\eqref{eq:3d_quad_dynamics} using an explicit Runge-Kutta method of 4th order with a step size of $\SI{0.5}{\milli\second}$.
We assume to have access to perfect odometry measurements of the quadrotor, ideal tracking of the commanded single-rotor thrusts, and that the MPC computation is instantaneous.
The simulator models drag effects caused both by the rotors as well as the fuselage. 
Additionally, zero-mean Gaussian noise forces and torques are simulated that act on the quadrotor body, as well as asymmetric noise on the motor voltage signals. 

In the simplified simulation, we investigate the influence of the number of training points of the GP on the predictive performance. 
As the choice of this hyperparameter constitutes a trade-off between model accuracy and computation time, we seek the model with the minimum number of inducing points that surpasses a desired performance threshold. 
This effect is investigated with two experiments: first, we analyze the trade-off between GP performance and optimization time. In the second experiment, we extend the comparison of different-sized GPs to closed-loop experiments on the three test trajectories.

Having identified the optimal size of the GP, we perform an additional set of experiments to compare the tracking performance on the circle and lemniscate test trajectories between the Nominal and RDRv baselines and our approach. 
Both trajectories are executed up to varying maximum speeds, where the highest speed pushes the platform to its physical limits.
The training set for both the RDRv and the GP models in this simulator is collected by executing random polynomial trajectories of high aggressiveness (such as Fig.~\ref{fig:trajectory_plots} left) with the quadrotor, up to $16\si{\meter\per\second}$ axis-wise.
This technique works well in simulation since it allows to explore densely all the ranges of operating points without risk of breaking the aircraft if tracking fails.

The results of the GP size analysis are summarized in Fig.~\ref{fig:gp_training_samples_tradeoff}. 
As can be seen, the complexity of the optimization problem approximately follows a linear function with respect to the number of training points.
Since the predictive performance barely increases when adding more than 20 inducing points, we identify the optimal range to be between 15 to 25 samples, corresponding to $4-5\:\si{\milli\second}$ of optimization time.
For comparative purposes, we illustrate in Fig.~\ref{fig:experiment_1_custom_sim} how two of our GP models perform in closed-loop tracking for 15 and 100 training samples.
It can be verified that a larger number of samples is strictly beneficial, but comes at the expense of an increase in optimization time.
In fact, such a large model is not usable for a real time application of our pipeline.
Based on this evidence, we chose to use \rebuttal{20} inducing points for the rest of this work.

\begin{table}[b]
\centering
\caption{Comparison of closed-loop tracking errors on the circle and lemniscate trajectories in simulation.}
\label{tab:rdrv_vs_gp_simulation}
\begin{tabular}{cc|c|c|cc|cc|}
\cline{3-8}
 &  & \multicolumn{6}{c|}{\textbf{Model}} \\ \cline{3-8} 
 &  & \textbf{Ideal} & \textbf{Nominal} & \multicolumn{2}{c|}{\textbf{RDRv}} & \multicolumn{2}{c|}{\textbf{GP-MPC}} \\ \hline %
\multicolumn{1}{|c|}{\rotatebox[origin=c]{90}{\textbf{Ref.}}} & \begin{tabular}[c]{@{}c@{}}$\mathbf{v_{peak}}$\\ $[\si{\meter\per\second}]$\end{tabular} & \multicolumn{1}{c|}{\begin{tabular}[c]{@{}c@{}}RMSE\\ $[\si{\milli\meter}]$\end{tabular}} & \begin{tabular}[c]{@{}c@{}}RMSE\\ $[\si{\milli\meter}]$\end{tabular} & \begin{tabular}[c]{@{}c@{}}RMSE\\ $[\si{\milli\meter}]$\end{tabular} & \cellcolor[HTML]{EFEFEF}\textbf{\%}$\downarrow$ & \begin{tabular}[c]{@{}c@{}}RMSE\\ $[\si{\milli\meter}]$\end{tabular} & \cellcolor[HTML]{EFEFEF}\textbf{\%}$\downarrow$ \\ \hline
\multicolumn{1}{|c|}{} & 4 & 0.1 & 114.1 & 18.1 & \cellcolor[HTML]{EFEFEF}84 & 16.1 & \cellcolor[HTML]{EFEFEF}\textbf{85} \\
\multicolumn{1}{|c|}{} & 8 & 0.4 & 241.2 & 56.9 & \cellcolor[HTML]{EFEFEF}76 & 25.4 & \cellcolor[HTML]{EFEFEF}\textbf{89} \\
\multicolumn{1}{|c|}{\multirow{-3}{*}{\rotatebox[origin=c]{90}{\textbf{Circle}}}} & 12 & 1.2 & 338.3 & 93.0 & \cellcolor[HTML]{EFEFEF}72 & 28.4 & \cellcolor[HTML]{EFEFEF}\textbf{93} \\ \hline
\multicolumn{1}{|c|}{} & 4 & 0.3 & 104.0 & 15.5 & \cellcolor[HTML]{EFEFEF}\textbf{85} & 16.3 & \cellcolor[HTML]{EFEFEF}84 \\
\multicolumn{1}{|c|}{} & 8 & 1.5 & 157.7 & 32.3 & \cellcolor[HTML]{EFEFEF}79 & 20.3 & \cellcolor[HTML]{EFEFEF}\textbf{87} \\
\multicolumn{1}{|c|}{\multirow{-3}{*}{\rotatebox[origin=c]{90}{\textbf{Lemn.}}}} & 12 & 4.2 & 212.4 & 60.6 & \cellcolor[HTML]{EFEFEF}71 & 24.4 & \cellcolor[HTML]{EFEFEF}\textbf{88} \\ \hline
\multicolumn{1}{c|}{} & \begin{tabular}[c]{@{}c@{}}\textbf{Opt. dt}\\ 
$[\si{\milli\second}]$\end{tabular} & \multicolumn{2}{c|}{1.32} & \multicolumn{2}{c|}{1.76} & \multicolumn{2}{c|}{4.13} \\ \cline{2-8} 
\end{tabular}
\end{table}

The main results of the closed-loop tracking experiments are summarized in Table~\ref{tab:rdrv_vs_gp_simulation}.
The table reports position tracking error in millimeters for both maneuvers at varying maximum speeds. 
While the \textit{Ideal} column indicates the tracking error in case of no unmodelled disturbances (i.e. the MPC dynamics model perfectly fit the actual system), the \textit{Nominal} column represents the baseline when no model augmentations are enabled.
Note that even though the MPC controller performs very well in the Ideal scenario, it does not achieve zero tracking error due to discretization effects.
Both the RDRv baseline as well as the GP-MPC significantly improve the tracking error compared to the non-augmented Nominal case.
However, while RDRv performs comparably to our approach up to speeds of $\SI{4}{\meter\per\second}$, it starts to fail for higher speeds due to its inability to model higher-order aerodynamic effects such as body drag. 
Our approach also captures these effects very well and shows consistent improvement for the full range of tested speeds.

\textbf{Gazebo Simulation} 
To verify the results obtained in the simplified simulation in a well-known quadrotor simulator, we also perform closed-loop tracking experiments in Gazebo~\cite{koenig2004design}.
We employ the AscTec Hummingbird quadrotor model using the RotorS extension~\cite{Furrer2016_RotorS}.
To properly evaluate the performance of our pipeline, we also use ground truth odometry measurements instead of a state estimator.
We collect a dataset containing velocities in the range $[-12, 12]\:\si{\meter\per\second}$ for training our models.
This dataset is obtained by tracking randomly generated aggressive trajectories, as with the simplified simulator.
We execute the circle and lemniscate trajectories at increasing speeds and compare the tracking performance of the Nominal and RDRv baselines, as well as our approach.
Note that once again we use completely independent training and test sets in our setup to ensure our models can generalize to new trajectories.

\begin{figure} %
\begin{adjustbox}{width=0.95\columnwidth}%
\begin{tikzpicture}

\definecolor{color2}{rgb}{0.549, 0.337, 0.294}
\definecolor{color0}{rgb}{1,0.498039215686275,0.0549019607843137}
\definecolor{color1}{rgb}{0.172549019607843,0.627450980392157,0.172549019607843}

\begin{groupplot}[group style={group size=1 by 2},height=4.2cm,width=11cm]
\nextgroupplot[
tick align=outside,
tick pos=left,
title={RMSE [m]},
x grid style={white!69.0196078431373!black},
xmajorgrids,
xmin=2, xmax=10.375,
xtick style={color=black},
y grid style={white!69.0196078431373!black},
ymajorgrids,
ylabel=Circle,
ymin=0.00637592751321536, ymax=0.223236553584884,
ytick style={color=black},
ytick={0,0.1,0.2,0.3},
yticklabels={0.0,0.1,0.2,0.3}
]
\addplot [semithick, color0, dashed, mark=*, mark size=3, mark options={solid}]
table {%
2.5 0.0590211587829369
4 0.0857927844485661
6 0.108559624342375
8 0.117163742002034
10 0.213379252399808
};
\addplot [semithick, color1, dashed, mark=*, mark size=3, mark options={solid}]
table {%
2.5 0.0165609953877652
4 0.0221310277076545
6 0.0306404567791664
8 0.0404345688491554
10 0.0531573297625603
};
\addplot [semithick, color2, dashed, mark=*, mark size=3, mark options={solid}]
table {%
2.5 0.0162332286982912
4 0.0257063785933315
6 0.0401032046787036
8 0.0562108116629014
10 0.0745087008841073
};

\nextgroupplot[
tick align=outside,
tick pos=left,
x grid style={white!69.0196078431373!black},
xlabel={max vel [m/s]},
xmajorgrids,
xmin=3, xmax=12,
xtick style={color=black},
y grid style={white!69.0196078431373!black},
ymajorgrids,
ylabel=Lemniscate,
ymin=0.00463646788873304, ymax=0.195135155792908,
ytick style={color=black},
ytick={0,0.05,0.1,0.15,0.2},
yticklabels={0.00,0.05,0.10,0.15,0.20}
]
\addplot [semithick, color0, dashed, mark=*, mark size=3, mark options={solid}]
table {%
3.4 0.0599632988933088
5.2 0.0896628378247211
8.8 0.117863472456754
11.6 0.137923120088675
};\label{gazebo_exp_1:plot1}
\addplot [semithick, color1, dashed, mark=*, mark size=3, mark options={solid}]
table {%
3.4 0.0132954991571046
5.2 0.0201875876726585
8.8 0.0311780110842474
11.6 0.0474231136226402
};\label{gazebo_exp_1:plot2}
\addplot [semithick, color2, dashed, mark=*, mark size=3, mark options={solid}]
table {%
3.4 0.0137892545827068
5.2 0.0225233716393806
8.8 0.0339684479861096
11.6 0.048280674216374
};\label{gazebo_exp_1:plot3}
\coordinate (bot) at (rel axis cs:1,0);%

\end{groupplot}

\path (top|-current bounding box.north)--
      coordinate(legendpos)
      (bot|-current bounding box.north);
\matrix[
    matrix of nodes,
    anchor=south,
    draw,
    inner sep=0.2em,
    draw
  ]at([yshift=1ex]legendpos)
  {
    \ref{gazebo_exp_1:plot1}& Nominal & [5pt]
    \ref{gazebo_exp_1:plot2}& GP-MPC 15 & [5pt]
    \ref{gazebo_exp_1:plot3}& RDRv 
    \\};
    
\end{tikzpicture}
\end{adjustbox}
\caption{Closed-loop position tracking error as a function of maximum velocity achieved in the RotorS Gazebo simulator in the circle and the lemniscate trajectories.}
\label{fig:experiment_1_gazebo}
\end{figure}
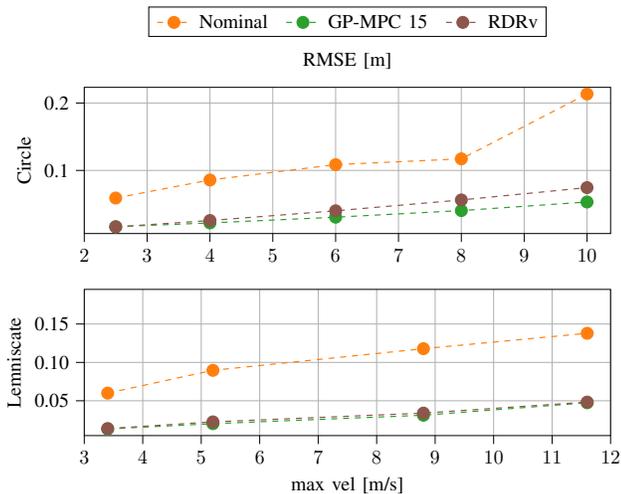

The main results of the Gazebo experiments are summarized in Fig.~\ref{fig:experiment_1_gazebo}.
In this case, both RDRv as well as our GP-MPC achieve very similar performance over the full range of tested speeds. 
This is expected, as the RotorS package implementation only simulates rotor drag as aerodynamic effect \cite{Furrer2016_RotorS}, which follows a linear mapping with respect to the body frame velocity~\cite{martin2010true}.
The true aerodynamic effects acting on a quadrotor however are a combination of rotor drag, body drag and turbulent effects caused by the propellers. 
We are analyzing these effects in more detail in the following section.

\subsection{Experiments in the Real World}
\label{sec:experiments_real_world}
Lastly, we compare the performance of our GP-MPC against both Nominal as well as RDRv controllers on a real quadrotor. 
We use a custom quadrotor that weighs $0.8\si{\kilogram}$ and has a thrust-to-weight ratio of 5:1. 
We run the controller on a laptop computer and send control commands in the form of collective thrust and desired bodyrates \rebuttal{at $\SI{50}{\hertz}$} to the quadrotor through a Laird RM024 radio module. %
A PID controller running onboard the quadrotor tracks the sent commands.
The quadrotor flies in an indoor arena equipped with an optical tracking system that provides pose estimates at $\SI{100}{\hertz}$.
Note that our control method also works with state estimates that are obtained differently than with a motion capture system. 
As in the simulation experiments, we compare the tracking error along both circle and lemniscate trajectories with speeds up to $\SI{14}{\meter\per\second}$. 

\begin{figure}[t]
    \centering
    \includegraphics[trim=0 150 0 150,clip,width=1.0\linewidth]{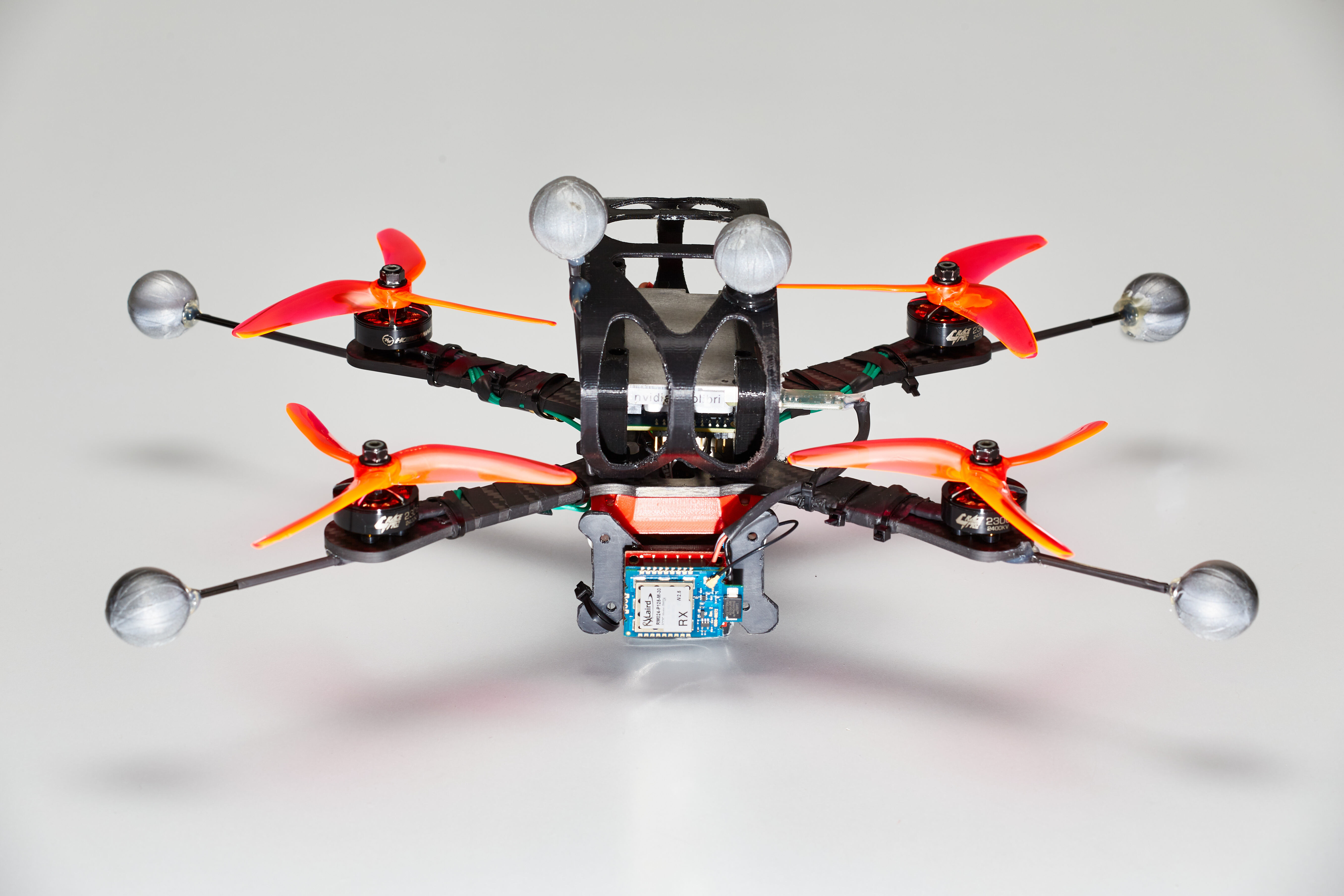}
    \caption{The quadrotor used for real-world experiments.}
    \label{fig:real_platform}
\end{figure}

\begin{figure} %
\begin{adjustbox}{width=\columnwidth}
\centerline{\includegraphics[width=0.95\columnwidth]{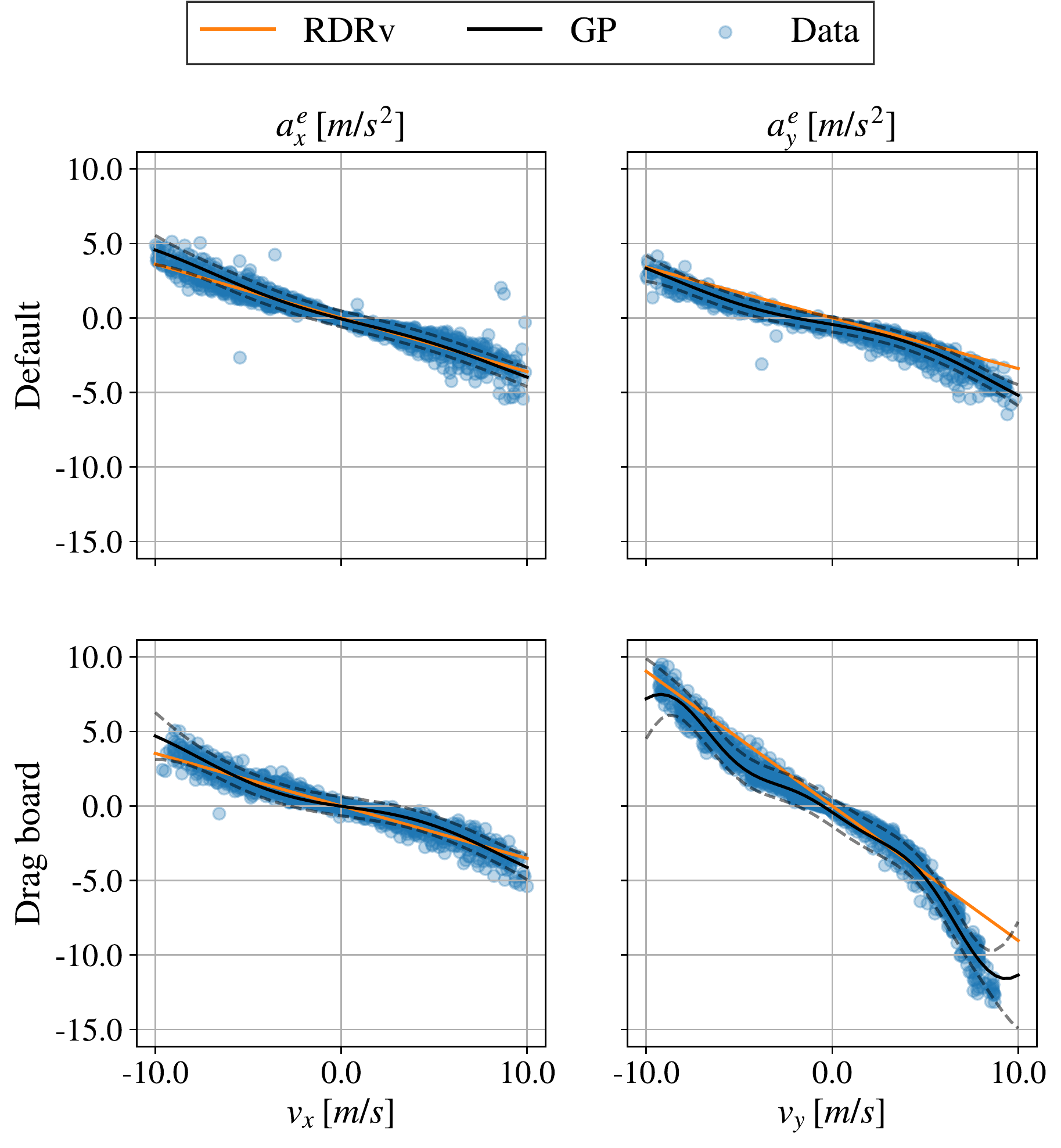}}
\end{adjustbox}
\caption{Aerodynamic effects observed in the real quadrotor platform along body axes $x$ (left column) and $y$ (right column) as a function of body frame velocity. The platform was studied in its default configuration (upper row), and with an additional flat board attached along the body x axis (lower row), resulting in a significantly increased body drag effect in the $y$ direction.}
\label{fig:rdrv_vs_gp_fit}
\end{figure}

To demonstrate that our approach can correct for complex aerodynamic effects, we perform the real world experiments in two settings: 
in setting \rom{1} we perform all maneuvers with the quadrotor as pictured in Fig.~\ref{fig:real_platform}, 
while in setting \rom{2} we extend the quadrotor body with a vertical drag board.
This drag board introduces additional asymmetric aerodynamic disturbance as can be seen in Fig.~\ref{fig:rdrv_vs_gp_fit}. 
\rebuttal{For the real world experiment, we use 20 inducing points on our GP's.}

\begin{table}[b]
\centering
\caption{Comparison of the RDRv and GP-MPC methods in the real world experiments.}
\label{tab:real_world_experiments}
\begin{tabular}{c|c|cc|cc|cc|}
\cline{2-8}
\textbf{} & \multicolumn{7}{c|}{\textbf{Model RMSE} $[\si{\milli\meter}]$} \\ \hline
\multicolumn{1}{|c|}{} &  & \textbf{GP} & \cellcolor[HTML]{EFEFEF} & \textbf{GP} & \cellcolor[HTML]{EFEFEF} &  & \cellcolor[HTML]{EFEFEF} \\
\multicolumn{1}{|c|}{\multirow{-2}{*}{\textbf{Ref.}}} & \multirow{-2}{*}{\textbf{Nomin.}} & \multicolumn{1}{l}{(circle)} & \multirow{-2}{*}{\cellcolor[HTML]{EFEFEF}\textbf{\%}$\downarrow$} & \multicolumn{1}{l}{(lemn.)} & \multirow{-2}{*}{\cellcolor[HTML]{EFEFEF}\textbf{\%}$\downarrow$} & \multirow{-2}{*}{\textbf{RDRv}} & \multirow{-2}{*}{\cellcolor[HTML]{EFEFEF}\textbf{\%}$\downarrow$} \\ \hline
\multicolumn{1}{|c|}{\textbf{Circle}} & 319.7 & 172.9 & \cellcolor[HTML]{EFEFEF}46 & 141.0 & \cellcolor[HTML]{EFEFEF}\textbf{56} & 168.3 & \cellcolor[HTML]{EFEFEF}47 \\ \hline
\multicolumn{1}{|c|}{\textbf{Lemn.}} & 396.2 & 254.2 & \cellcolor[HTML]{EFEFEF}\textbf{36} & 266.3 & \cellcolor[HTML]{EFEFEF}33 & 269.3 & \cellcolor[HTML]{EFEFEF}33 \\ \hline
\end{tabular}
\end{table}

\begin{table}[b]
\centering
\caption{Velocity-dependent tracking performance of the augmented MPC methods on the circle trajectory.}
\label{tab:circle_histogram_error}
\begin{tabular}{cc|c|cc|cc|}
\cline{3-7}
 &  & \multicolumn{5}{c|}{\textbf{Model}} \\ \cline{3-7} 
\textbf{} & \textbf{} & \textbf{Nominal} & \multicolumn{2}{c|}{\textbf{RDRv}} & \multicolumn{2}{c|}{\textbf{GP-MPC 20}} \\ \hline
\multicolumn{1}{|c|}{\textbf{Config.}} & \begin{tabular}[c]{@{}c@{}}$\mathbf{v_{range}}$\\ $[\si{\meter\per\second}]$\end{tabular} & \begin{tabular}[c]{@{}c@{}}RMSE \\ $[\si{\meter}]$\end{tabular} & \begin{tabular}[c]{@{}c@{}}RMSE\\ $[\si{\meter}]$\end{tabular} & \cellcolor[HTML]{EFEFEF}\textbf{\%}$\downarrow$ & \begin{tabular}[c]{@{}c@{}}RMSE\\ $[\si{\meter}]$\end{tabular} & \cellcolor[HTML]{EFEFEF}\textbf{\%}$\downarrow$ \\ \hline
\multicolumn{1}{|c|}{} & 0-2 & 0.087 & 0.130 & \cellcolor[HTML]{EFEFEF}-49 & 0.109 & \cellcolor[HTML]{EFEFEF}-25 \\
\multicolumn{1}{|c|}{} & 2-4 & 0.233 & 0.119 & \cellcolor[HTML]{EFEFEF}49 & 0.103 & \cellcolor[HTML]{EFEFEF}\textbf{56} \\
\multicolumn{1}{|c|}{} & 4-6 & 0.329 & 0.177 & \cellcolor[HTML]{EFEFEF}46 & 0.129 & \cellcolor[HTML]{EFEFEF}\textbf{61} \\
\multicolumn{1}{|c|}{} & 6-8 & 0.458 & 0.210 & \cellcolor[HTML]{EFEFEF}54 & 0.154 & \cellcolor[HTML]{EFEFEF}\textbf{66} \\
\multicolumn{1}{|c|}{\multirow{-5}{*}{\rotatebox[origin=c]{90}{\textbf{Default}}}} & 8-10 & 0.531 & 0.192 & \cellcolor[HTML]{EFEFEF}\textbf{64} & 0.203 & \cellcolor[HTML]{EFEFEF}62 \\ \hline
\multicolumn{1}{|c|}{} & 0-2 & 0.197 & 0.132 & \cellcolor[HTML]{EFEFEF}33 & 0.060 & \cellcolor[HTML]{EFEFEF}\textbf{69} \\
\multicolumn{1}{|c|}{} & 2-4 & 0.346 & 0.287 & \cellcolor[HTML]{EFEFEF}17 & 0.078 & \cellcolor[HTML]{EFEFEF}\textbf{77} \\
\multicolumn{1}{|c|}{} & 4-6 & 0.564 & 0.381 & \cellcolor[HTML]{EFEFEF}32 & 0.141 & \cellcolor[HTML]{EFEFEF}\textbf{75} \\
\multicolumn{1}{|c|}{} & 6-8 & 0.837 & 0.463 & \cellcolor[HTML]{EFEFEF}45 & 0.219 & \cellcolor[HTML]{EFEFEF}\textbf{74} \\
\multicolumn{1}{|c|}{\multirow{-5}{*}{\rotatebox[origin=c]{90}{\textbf{Drag board}}}} & 8-10 & 0.912 & \emph{Crash} & \cellcolor[HTML]{EFEFEF}\emph{?} & 0.379 & \cellcolor[HTML]{EFEFEF}\textbf{59} \\ \hline
\end{tabular}
\end{table}

Table~\ref{tab:real_world_experiments} summarizes the results of our real world experiments in setting~\rom{1}. 
We train two GP models on the circle and lemniscate trajectories, and use them at test time in all permutations.
As can be seen, our methods as well as the RDRv baseline improve tracking performance by up to $50$\%, with our approach slightly outperforming the RDRv baseline. 
This result can be explained by the fact that the quadrotor platform used in setting~\rom{1} is very compact and powerful, rendering the main source of disturbance being the rotor drag. 
Rotor drag is a linear effect, which can be well compensated for by the linear RDRv model augmentation. 
The slight improvement of \gpmpc\ over RDRv in this setting can be explained by the ability of the GPs to also account for imperfect thrust mappings. 

Finally, we compare the circle tracking for settings~\rom{1} and \rom{2} in Table~\ref{tab:circle_histogram_error}.
As can be seen, our approach significantly outperforms \textit{RDRv} in setting~\rom{2}, where the latter fails to capture the full nonlinearity of aerodynamic effects. 
This can be verified also in Fig.~\ref{fig:rdrv_vs_gp_fit}, where the linear fit leads to significant bias.

\section{Conclusion}\label{sec:conclusion}

In this work, we propose the usage of Gaussian Processes to augment the nominal dynamics of a quadrotor to compensate for aerodynamic effects. %
This GP-based model augmentation is integrated in a Model Predictive Controller and the resulting system significantly improves positional tracking error, both in simulation and on a real quadrotor.
Using data from previously recorded flights, the GP's are trained to predict the acceleration error of the nominal model given its current velocity in body frame.

In extensive experiments in simulation and the real world, we show that our approach outperforms a state-of-the-art linear drag model. 
\rebuttal{Furthermore, our GP-augmented controller opens up interesting lines of follow-up research for future development}. 
\rebuttal{On one hand}, we plan to make use of the predicted uncertainty to perform safe agile trajectories close to obstacles.
\rebuttal{On the other hand, leveraging the fast fitting time of our GP models (in the order of seconds), training and control loop can be executed in parallel on separate threads in real time during flight.
This would enable to adapt the dynamics model to varying external or internal conditions such as wind disturbance or battery voltage.}

\rebuttal{
\section{Acknowledgment}
The authors thank Leonard Bauersfeld for his help with the photographic illustrations.
}

\balance
{\footnotesize
\bibliographystyle{IEEEtran}
\bibliography{references}
}

\end{document}